\documentclass[runningheads]{llncs}

\usepackage{accv}

\usepackage{accvabbrv}

\usepackage{graphicx}
\usepackage{booktabs}
\usepackage{colortbl}
\usepackage{gensymb}
\usepackage{textcomp}
\usepackage{multirow}
\usepackage{multicol}
\usepackage{wrapfig}
\usepackage{rotating}
\usepackage{siunitx}
\usepackage[table]{xcolor} 
\usepackage{etoolbox}
\usepackage{multirow}
\usepackage{multicol}
\usepackage{pifont}

\newrobustcmd{\B}{\bfseries}    
\newrobustcmd{\U}{\underline}
\newcommand{\cmark}{\ding{51}}%
\newcommand{\xmark}{\ding{55}}%

\definecolor{lavender}{rgb}{0.9, 0.9, 0.98}
\definecolor{lightgray}{rgb}{0.5, 0.5, 0.5}
\definecolor{beaublue}{rgb}{0.74, 0.83, 0.9}
\usepackage[table]{xcolor} 
\usepackage{etoolbox}

\newcommand{\PAR}[1]{\vskip4pt \noindent{\bf #1~}}

\makeatletter
\newcommand{\@cellbg}[1]{\gdef\currentcellbg{\cellcolor{#1}}}
\makeatother

\definecolor{pastel_yellow}{RGB}{255, 243, 176} 
\definecolor{pastel_blue}{RGB}{190, 225, 245}   
\definecolor{pastel_pink}{RGB}{244, 210, 194}   
\newcommand{\M}[1]{\mathbf{#1}}

\newrobustcmd{\Best}{\cellcolor{pastel_yellow}}
\newrobustcmd{\Second}{\cellcolor{pastel_blue}}
\newrobustcmd{\Third}{\cellcolor{pastel_pink}}

\usepackage[accsupp]{axessibility}  

\usepackage{hyperref}

\usepackage{orcidlink}

\begin{document}

\title{DGSfM: Depth-Guided Scale-Aware \texorpdfstring{\\}{ }Global Structure-from-Motion} 

\titlerunning{DGSfM}

\author{Sithu Aung\inst{1}\orcidlink{0009-0006-4151-8545}\and
Viktor Kocur\inst{2}\orcidlink{0000-0001-8752-2685}\and
Yaqing Ding\inst{3}\orcidlink{0000-0002-7448-6686} \and \\
Torsten Sattler\inst{4}\orcidlink{0000-0001-9760-4553}\and
Zuzana Kukelova\inst{1}\orcidlink{0000-0002-1916-8829}
}

\authorrunning{S. Aung et al.}

\institute{
VRG, FEE, Czech Technical University in Prague \and
FMPH, Comenius University in Bratislava \and
Southeast University \and
CIIRC, Czech Technical University in Prague 
}

\maketitle

\begin{abstract}
  Global Structure-from-Motion (SfM) is an efficient paradigm for recovering camera poses and sparse 3D structure from unordered images.  
  However, its reliance on scale-ambiguous epipolar geometry makes global positioning sensitive to noisy baseline estimates and weak view-graph constraints, while false edges from visually ambiguous pairs can further degrade reconstruction.
  We propose DGSfM, a depth-aware global SfM pipeline that uses monocular depth maps as a scalable prior while preserving explicit multi-view optimization.
  For each image pair, we use a depth-aware relative pose solver to convert scale-ambiguous epipolar constraints into scale-aware relative pose constraints.
  We further improve robustness through view-graph filtering and depth-consistency-based correspondence pruning, which suppress false edges and matches that remain plausible under epipolar geometry alone.
  Finally, global scale averaging and depth-guided pose-point initialization align monocular depth maps into a common reconstruction scale and provide stable initialization for global positioning and bundle adjustment.
  Experiments on ETH3D and IMC2021 show that DGSfM consistently improves over strong global SfM baselines across sparse and dense matching front-ends, achieving substantial gains in pose accuracy. 
  Code and project page: \href{https://dgsfm.github.io/}{dgsfm.github.io}.
  \keywords{Structure-from-Motion \and Global SfM \and Monocular Depth}

\end{abstract}

\section{Introduction}
\label{sec:intro}

\begin{figure*}[t]
  \centering
\includegraphics[width=\textwidth]{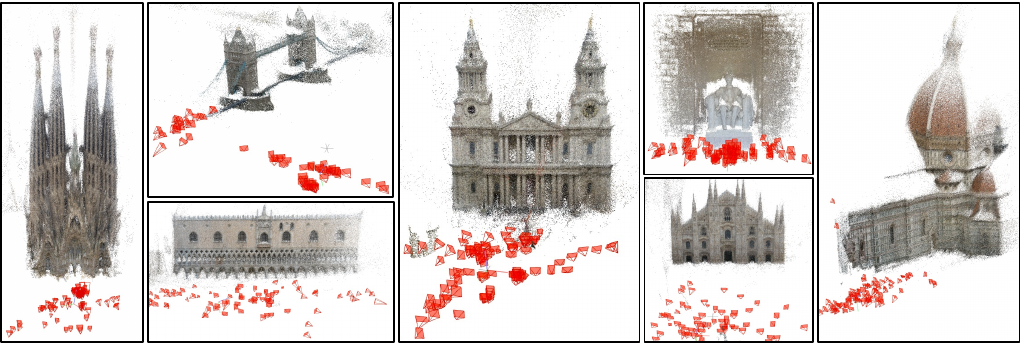}
  \caption{Reconstruction results of DGSfM using RoMa~\cite{romav1} matches on the full image sets from the IMC2021~\cite{imc2021} dataset.
  DGSfM integrates monocular depth into global SfM to improve scale estimation, view-graph robustness, and pose-point initialization, producing globally consistent camera poses and sparse/semi-dense 3D structure.
  }
  \label{fig:teaser}
\end{figure*}

Structure-from-Motion (SfM) is a core tool for recovering camera poses and sparse 3D structure from images, with applications in 3D reconstruction, localization, visual mapping, and spatial data generation.
Among SfM paradigms~\cite{wu2013towards, snavely2006photo, moulon2016openmvg, agarwal2011building}, global SfM~\cite{crandall2011discrete, wilson2014robust} is particularly attractive because it estimates camera poses jointly from a view graph, avoiding the sequential registration order and repeated local optimization of incremental pipelines.
Recent works such as GLOMAP~\cite{glomap} show that global SfM can be both scalable and accurate when the input view graph is reliable.
However, global SfM still largely relies on pairwise epipolar geometry, where relative translations are known only up to scale and where visually plausible but incorrect image pairs can survive geometric verification.
This makes global positioning vulnerable to noisy baseline estimates, ambiguous matches, and inconsistent local geometry, especially in unordered image collections with repeated structures, weak overlap, or limited parallax.

Recent learned methods provide powerful geometric cues, but they do not by themselves remove the need for explicit multi-view optimization.
Dense matchers~\cite{romav1, romav2, zhang2025ufm} and detector-free SfM methods~\cite{dfsfm,densesfm,mvroma} improve correspondence coverage and robustness in low-texture or wide-baseline settings, yet the subsequent SfM optimization often remains governed by scale-ambiguous relative poses and track consistency.
Feed-forward reconstruction models~\cite{vggsfm,vggt,pi3,lin2025depthanything3, keetha2026mapanything} can predict cameras and dense structure directly from images, but their accuracy, scalability, and global consistency may degrade for large unordered collections or scenes outside their training distribution compared to classical feature-based SfM methods~\cite{meza2026benchmarking, pan2026global}.
Monocular depth estimators~\cite{depth_anything_v2,depthpro,wang2025moge,wang2025moge2, hu2024metric3dv2} offer another complementary source of information: they provide per-image geometry, often with metric or near-metric scale, that can regularize degenerate multi-view configurations.
The key question is therefore not whether learned geometry should replace SfM, but how it can be injected into global SfM so that scale, filtering, and initialization become more reliable while final reconstruction remains constrained by multi-view geometry.

We propose DGSfM, a depth-aware global SfM framework that integrates monocular depth estimates into the main stages of global reconstruction.
Our starting point is the observation that monocular depth can turn the pairwise constraints used by global SfM from scale-ambiguous relations into scale-aware geometric constraints.
For each verified image pair, we use a depth-aware relative pose solver~\cite{ding2025reposed} to estimate the relative rotation, the scaled relative translation, the relative depth-scale factor, and camera intrinsics.
These pairwise estimates provide stronger input for the global pipeline than epipolar geometry alone, because they expose the metric compatibility between two views before global positioning.
At the same time, we do not trust monocular predictions as final geometry; instead, we use them as priors that guide robust filtering and initialization before standard multi-view optimization refines poses and points.

DGSfM improves robustness by filtering both the view graph and the matches using scale-aware depth consistency.
At the graph level, we combine pairwise visual disambiguation~\cite{doppelgangers++} with triplet-guided edge pruning~\cite{manam2024leveraging} to remove image pairs that are visually similar or locally plausible but inconsistent with neighboring views.
At the correspondence level, we back-project matches with monocular depth, transfer them across the scaled relative pose, and retain only those that agree in both reprojection and target-view depth.
This rejects matches that may satisfy weak epipolar constraints but are incompatible with the estimated 3D geometry, leading to cleaner tracks for global positioning and bundle adjustment.

DGSfM further uses monocular geometry to initialize the global reconstruction in a common scale.
Since monocular depth predictions can contain image-wise scale drift, we first solve a robust global scale averaging problem over the filtered view graph.
The recovered image-wise scale factors align all depth maps into a shared reconstruction scale.
We then initialize camera centers by chaining scaled relative translations over a maximum spanning tree and initialize sparse 3D points by lifting multi-view track observations through the scaled depth maps.
These depth-guided camera and point estimates are passed to the GLOMAP-style global positioning stage~\cite{glomap}, followed by bundle adjustment, so the final solution is still determined by explicit multi-view consistency.

Experiments on ETH3D~\cite{eth3dsfm} and IMC2021~\cite{imc2021} show that the proposed depth-aware global SfM pipeline improves camera pose accuracy across multiple feature and matching front-ends.
With sparse features, DGSfM consistently improves over COLMAP~\cite{colmap} and GLOMAP~\cite{glomap}. With dense RoMa~\cite{romav1} matches, it achieves strong performance compared to dense SfM baselines~\cite{dfsfm,densesfm,mvroma}.
Some example reconstructions are shown in Fig.~\ref{fig:teaser}.
These results support the central design principle of our method: learned monocular geometry is most effective when used as a metric prior to strengthen scale estimation, outlier rejection, and initialization within a global SfM pipeline, while leaving the final reconstruction governed by explicit multi-view geometric optimization.
Our contributions are: 
\begin{itemize}
  \item We introduce a depth-aware global SfM pipeline that integrates learned monocular geometry estimates into relative pose estimation, filtering, scale alignment, and global initialization.
  \item We propose scale-aware view-graph and correspondence filtering that removes visually ambiguous edges and depth-inconsistent matches before global optimization.
  \item We develop global scale averaging and depth-guided pose-point initialization to provide stable metric initialization for GLOMAP-style global positioning and bundle adjustment.
  \item We show consistent improvements on ETH3D and IMC2021 across sparse and dense matching settings, showing that monocular geometry can robustly strengthen global SfM while preserving explicit multi-view optimization.
\end{itemize}

\section{Related Work}
\PAR{Traditional SfM.} 
Traditional Structure-from-Motion (SfM) pipelines~\cite{agarwal2011building, wu2013towards, snavely2006photo, colmap} recover camera poses and sparse 3D structure through feature matching~\cite{sift, superpoint, sarlin2020superglue}, geometric verification~\cite{colmap}, triangulation~\cite{hartley1997triangulation}, and bundle adjustment~\cite{triggs1999bundle}. 
Incremental systems such as COLMAP~\cite{colmap, schonberger2016pixelwise} are widely used due to their robustness and accuracy, but they can be computationally expensive because cameras are registered sequentially and repeatedly optimized through bundle adjustment. 
Global SfM methods~\cite{gtsfm, moulon2016openmvg, wilson2014robust, Cui_2015_ICCV} improve scalability by estimating camera poses jointly from a view graph. 
Recent systems such as GLOMAP~\cite{glomap} show that global reconstruction can achieve strong efficiency while maintaining competitive accuracy.
However, most global SfM methods still rely mainly on epipolar geometry~\cite{hartley2003multiple}, where pairwise translation is recovered only up to an unknown scale. 
As a result, global positioning and point initialization depend on noisy track constraints and indirect scale estimation. 
Moreover, outlier image pairs and inconsistent correspondences can strongly affect the global solution. 
In contrast, our work improves global SfM by introducing monocular geometry into the reconstruction process, reducing scale ambiguity and improving the reliability of pose and structure initialization.

\PAR{Feature Refinement and Dense Matching-based SfM.} 
Recent works improve SfM by refining sparse features or replacing sparse keypoint matching~\cite{sarlin2020superglue, lindenberger2023lightglue} with dense and detector-free correspondences~\cite{sun2021loftr, zhang2025ufm, romav1, romav2}. 
PixSfM~\cite{pixsfm} introduces featuremetric refinement, where keypoint locations, camera poses, and 3D points are refined by aligning learned dense features across multiple views. 
Detector-Free SfM~\cite{dfsfm} further explores the use of detector-free matchers for SfM. 
It first obtains dense or semi-dense matches and then constructs multi-view tracks for reconstruction.
Dense-SfM~\cite{densesfm} improves track consistency through multi-view refinement, and Gaussian splatting-based track extension. 
MV-RoMa~\cite{mvroma} extends this direction by transforming pairwise dense matching into multi-view track reconstruction, jointly refining correspondences from one source view to multiple co-visible target views. 
These methods show that dense and multi-view-aware correspondences can significantly improve reconstruction density and accuracy. 
And they focus on improving feature matching and track generation, without modifying the rest of the (global) SfM pipeline.  
Instead of designing a new multi-view matcher or refining dense tracks with an additional correspondence model, we focus on improving global SfM using monocular geometry estimates. 
Our work is thus complementary to these methods.

\PAR{Feed-forward 3D Reconstruction.} 
Recent feed-forward 3D reconstruction methods aim to recover 3D geometry and camera information directly from images using large neural networks. 
DUSt3R~\cite{dust3r} predicts dense point maps and enables relative pose estimation and reconstruction without a traditional SfM pipeline. 
Follow-up methods such as MASt3R~\cite{mast3r}, VGGT~\cite{vggt}, Pi3~\cite{pi3}, DepthAnything3~\cite{lin2025depthanything3}, and VGGT-Omega~\cite{wang2026vggtomega} further improve dense correspondence, camera prediction, and multi-view reconstruction. 
These methods demonstrate that strong learned geometric priors can work in cases where sparse feature matching is difficult.
Despite their impressive progress, feed-forward methods may still lack the accuracy~\cite{lin2025depthanything3, wang2026vggtomega}, scalability, and global consistency of optimization-based SfM, especially for large unordered image collections~\cite{deng2025vggtlong, maggio2026vggtslam, wang20253d}. 
They also often depend on learned priors that may not generalize perfectly to unseen scenes. 
Recently, GLUEMAP~\cite{pan2026global} combined feed-forward reconstruction and a global SfM pipeline, using the learned local geometry while preserving the scalability and global consistency of optimization-based reconstruction.
Our work is complementary but differs in focus: rather than relying on a feed-forward reconstruction backbone, we use learned monocular geometry to strengthen a global SfM framework while preserving explicit multi-view geometric consistency.

\PAR{Monocular Geometry Priors for SfM.}
Monocular depth estimation has recently shown strong generalization across diverse scenes. 
Relative depth models such as MiDaS~\cite{Ranftl2022, birkl2023midas} and DPT~\cite{Ranftl2021}, as well as recent foundation models such as Depth Anything V1/V2~\cite{depth_anything_v1, depth_anything_v2}, MoGe1~\cite{wang2025moge}, provide robust single-image geometric cues. 
Metric depth models, including Metric3D~\cite{yin2023metric3d, hu2024metric3dv2}, UniDepth~\cite{piccinelli2024unidepth, piccinelli2025unidepthv2}, DepthPro~\cite{depthpro}, and MoGe2~\cite{wang2025moge2}, further aim to recover depth in real-world scale.
However, monocular depth alone remains ambiguous and suffers from scale inconsistency, domain shift, and local prediction errors.
Recent relative pose solvers such as RePoseD~\cite{ding2025reposed} and MADPose~\cite{yu2025relative} incorporate mono-depth into two-view geometry, enabling scale-aware relative pose estimation and correction of depth-scale errors.
Another line of work uses monocular geometry as a prior within SfM, where multi-view constraints can correct single-view errors.
MP-SfM~\cite{mpsfm} is the closest work, incorporating mono-depth and surface normal priors into an incremental SfM pipeline for low-overlap, low-parallax, and highly symmetric scenes.
However, it assumes known camera intrinsics and mainly uses monocular priors as local geometric regularization.
In contrast, our work focuses on global SfM.
We use mono-depth to address scale ambiguity and instability in global reconstruction through depth-aware relative pose estimation, scale averaging, and initialization.
Our pipeline also estimates intrinsics through focal averaging from depth-aware relative poses, reducing the need for known calibration. 
MASt3R-SfM~\cite{mast3rsfm} is similar to our approach as both first estimate relative poses and depth maps, compute consistent depth map scaling factors, align pairwise reconstructions, and finally refine the initial pose estimates. 
MASt3R-SfM uses a feedforward network to jointly predict the relative pose and the depth maps, while our approach uses depth maps to robustify the relative pose estimation process. 
MASt3R-SfM relies on estimating similarity transformations to obtain an initial set of poses whereas our approach uses rotation averaging and positioning. 
Our results show that our approach clearly outperforms MASt3R-SfM. 


\section{Method}
\label{sec:method}
\begin{figure*}[t]
  \centering
\includegraphics[width=\textwidth]{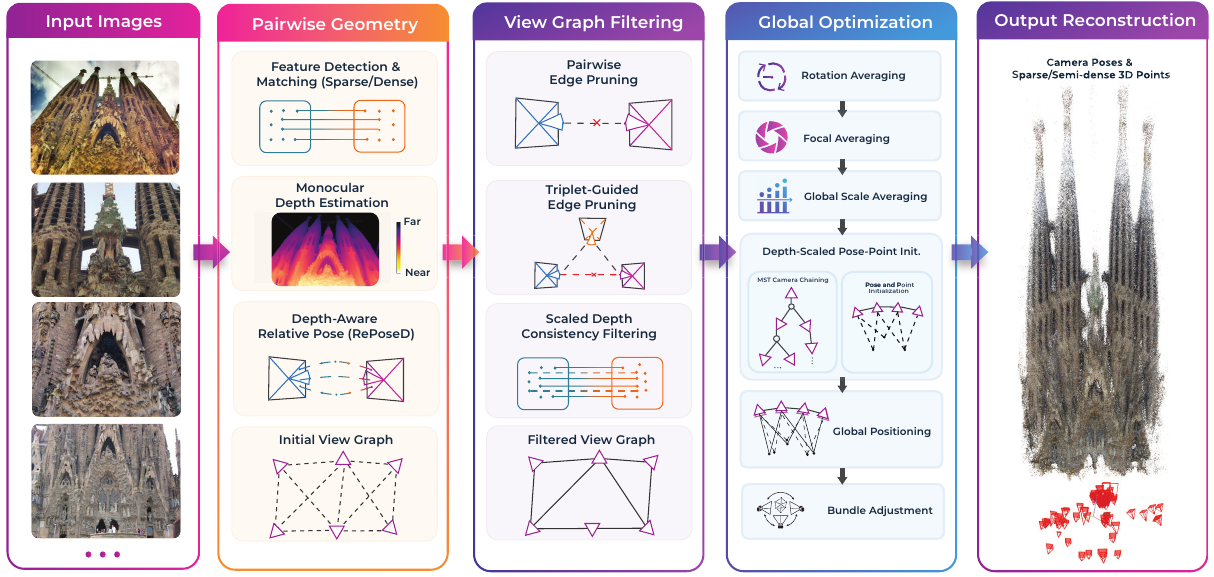}
  \caption{Overview of DGSfM. Given unordered images, we extract sparse or dense matches, monocular depth, and compute depth-aware relative poses and pairwise focals to construct an initial view graph, which is refined through pairwise and triplet-guided edge pruning, and scaled depth consistency checks. Mapping is done with rotation averaging, focal and scale averaging, global positioning with depth-scaled initialization, and bundle adjustment, producing camera poses and sparse or semi-dense 3D points.
  }
  \label{fig:pipeline}
\end{figure*}

Given an unordered image collection $\mathcal{I}=I_{i=1}^{N}$, our goal is to recover globally consistent camera poses $\{(\mathbf{R}_i,\mathbf{t}_i)\}_{i=1}^{N}$, sparse 3D scene points ${\mathbf{X}_p}$, and camera calibration parameters ${\mathbf{K}_i}$.
We build upon the global SfM paradigm, specifically GLOMAP~\cite{glomap}, where an initial view graph is constructed from geometrically verified image pairs, followed by global rotation averaging, global positioning, and finally bundle adjustment.
In contrast to conventional global SfM pipelines, which mainly rely on scale-ambiguous epipolar geometry, our method introduces monocular depth as a metric geometric prior.
This prior is used to improve relative pose estimation, filter unreliable view-graph edges, prune inconsistent correspondences, and initialize camera poses and 3D points in a depth-aware manner.
The overall pipeline is shown in Fig.~\ref{fig:pipeline}.

\subsection{Preliminaries}
\PAR{GLOMAP.}
Let $\mathcal{G}=(\mathcal{V},\mathcal{E})$ be the view graph, where each node $i\in\mathcal{V}$ represents an image and each edge ($(i,j)\in\mathcal{E}$) corresponds to a geometrically verified image pair~\cite{colmap}.
For each 
edge, two-view geometry~\cite{hartley2003multiple} provides a relative rotation $\mathbf{R}_{ij}$ and a translation direction $\bar{\mathbf{t}}_{ij}$.
GLOMAP~\cite{glomap} first estimates globally consistent camera rotations $\mathbf{R}_i$ by rotation averaging (RA)~\cite{chatterjee2013efficient}, such that $\mathbf{R}_{ij} = \mathbf{R}_j\mathbf{R}_i^{\top}$.
After RA, 
classical global SfM pipelines~\cite{gtsfm, moulon2016openmvg, Cui_2015_ICCV} estimate camera centers through translation averaging~\cite{zhuang2018baseline, wilson2014robust}. 
However, the translation recovered from epipolar geometry is only known up to scale, which makes translation averaging sensitive to noisy translation direction estimates and outlier edges. 
GLOMAP avoids this separate translation averaging stage by directly solving a global positioning problem, where camera centers and 3D points are estimated jointly.
While this avoids explicit translation averaging, the optimization can converge to slightly different solutions across different runs.
In our method, we keep the same GLOMAP-style global positioning objective, but use a deterministic depth-guided initialization derived from scaled monocular depth and scale-aware pairwise geometry.

\PAR{Depth-aware Relative Pose Solver.}
The relative translation recovered from epipolar geometry~\cite{hartley2003multiple} is defined only up to an unknown scale. 
To reduce this ambiguity, we estimate a scaled relative pose for each image pair using monocular depth.
For each image $I_i$, a monocular depth predictor~\cite{wang2025moge, wang2025moge2} provides a depth map $D_i$, which may be metric or defined up to an unknown scale.
Given feature correspondences between $I_i$ and $I_j$, we follow the depth-aware relative pose formulation of RePoseD~\cite{ding2025reposed} to estimate:
\begin{equation}
\mathbf{T}_{ij} = 
[
(\mathbf{R}_{ij}, \mathbf{t}_{ij}),
s_{ij}, \mathbf{K}_i, \mathbf{K}_j
] \enspace,
\label{eqn:solver}
\end{equation}
where ($\mathbf{R}_{ij}$, $\mathbf{t}_{ij}$) is the relative rotation and scaled relative translation, and $(\mathbf{K}_i, \mathbf{K}_j)$ are intrinsics estimated from the solver. Let $s_i$ denote the global depth scale of image $i$, $s_{ij} = {s_j}/{s_i}$ denotes the relative depth-scale factor between the two views. The scaled relative translation is defined w.r.t. $s_i$, \ie, $\M t_{ij} = \M t_{gt}/s_i$, where $\M t_{gt}$ is the translation with ground-truth scale. 
This converts each image pair from a scale-ambiguous epipolar constraint into a scale-aware relative pose constraint, providing stronger pairwise geometry for subsequent rotation averaging and global positioning.

\subsection{Robust View-Graph Filtering}
The robustness of global SfM depends critically on the quality of the input view graph.
False edges caused by repeated structures, symmetric facades, visually similar regions, or accidental feature matches can introduce inconsistent relative poses and severely degrade global pose estimation.
Although standard geometric verification removes many incorrect matches, visually ambiguous image pairs may still satisfy epipolar constraints and remain in the graph.
We therefore apply robust view-graph filtering before global optimization to remove unreliable image pairs and improve the consistency of the reconstruction.

\PAR{Pairwise and Triplet-Guided Edge Pruning.}
We first adopt pairwise visual disambiguation following Doppelgangers++~\cite{doppelgangers++}.
For each candidate edge $(i, j)$, the model predicts a confidence score $c_{ij}^{\mathrm{DG}}\in[0,1]$, which indicates whether the two images observe the same physical surface.
We retain an edge only if
$c_{ij}^{\mathrm{DG}} > \tau_{\mathrm{DG}}$,
where $\tau_{\mathrm{DG}}$ is a confidence threshold.
This step removes false edges caused by visual ambiguity before they influence global pose estimation.

We further exploit triplet consistency following camera-triplet-based SfM filtering~\cite{manam2024leveraging}.
Let $\mathcal{T}$ denote the set of valid triplets with edge $(i,j)$ as a member.
The support score of edge $(i,j)$ in a triplet $t \in \mathcal{T}$ can be computed as
\begin{equation}
q_{ij}^t
=
\frac{n_{ij}}{\max_{(k, l) \in t}n_{kl}} \enspace , 
\end{equation}
where $n_{ij}$ is the number of inliers of edge $(i,j)$.
The average score of each edge across all triplets in the view-graph can be computed as:
\begin{equation}
q_{ij}
=
\frac{\sum_{t\in\mathcal{T}}q_{ij}^t}{|\mathcal{T}|} \enspace .
\end{equation}
An edge is retained only if 
$q_{ij}>\tau_{\mathrm{tri}}$,
where $\tau_{\mathrm{tri}}$ is the triplet-support threshold~\cite{manam2024leveraging}.
This suppresses image pairs that appear plausible in isolation but are inconsistent with neighboring views.
The resulting filtered graph preserves well-supported connectivity while reducing 
triplet-inconsistent outlier edges.

\PAR{Scaled Depth Consistency Filtering.}
Even after geometric verification and view-graph filtering, a retained image pair may still contain incorrect feature correspondences.
Traditional SfM pipelines mainly reject such matches using epipolar error and  cheirality test.
However, these criteria may still accept ambiguous matches in repetitive regions or low-parallax image pairs.
Since our pipeline provides both scale-aware relative poses and monocular depth estimates, we further filter correspondences using cross-view scaled depth consistency.

Given a correspondence $(\mathbf{x}_i,\mathbf{x}_j)$ between images $I_i$ and $I_j$, we first back-project $\mathbf{x}_i$ into the coordinate frame of image $i$ using its monocular depth:
\begin{equation}
\mathbf{X}_i
=
\Pi_i^{-1}\bigl(\mathbf{x}_i, D_i(\mathbf{x}_i)\bigr) \enspace,
\end{equation}
where $\Pi_i^{-1}(\cdot)$ denotes back-projection with camera intrinsics $\mathbf{K}_i$ (estimated from Eqn.~\ref{eqn:solver}), and predicted monocular depth $D_i(\M x_i)$.
We then transform the point to the coordinate frame of image $j$ using the scaled relative pose $(\mathbf{R}_{ij},\mathbf{t}_{ij})$:
\begin{equation}
\tilde{\mathbf{X}}_j
=
\mathbf{R}_{ij}\mathbf{X}_i + \mathbf{t}_{ij} \enspace.
\end{equation}
A correspondence is considered reliable only if the transformed point is consistent with the matched location and depth in the target view.
We therefore compute the reprojection error
\begin{equation}
e_{ij}^{\pi}
=
\left\| 
\pi\bigl(\mathbf{K}_j, \tilde{\mathbf{X}}_j\bigr)
-
\mathbf{x}_j
\right\|_2 \enspace,
\end{equation}
and the scaled depth consistency error
\begin{equation}
e_{ij}^{d}
=
\frac{
\left|
[\tilde{\mathbf{X}}_j]_z
-
s_{ij}D_j(\mathbf{x}_j)
\right|
}{s_{ij}D_j(\mathbf{x}_j)} \enspace, 
\end{equation}
where $[\cdot]_z$ denotes the depth value in the target camera frame, and $s_{ij}$ is the relative depth-scale factor from Eqn.~\ref{eqn:solver}. 
The correspondence is retained only if
\begin{equation}
e_{ij}^{\pi}<\tau_{\pi}
\quad \text{and} \quad
e_{ij}^{d}<\tau_d \enspace,
\end{equation}
where $\tau_{\pi}$ and $\tau_d$ are reprojection and depth consistency error thresholds.
The same consistency check is also applied symmetrically from image $j$ to image $i$.
This filtering step removes matches that may satisfy 
epipolar constraints but are inconsistent with the estimated monocular geometry.
The remaining matches provide cleaner feature tracks for global positioning and bundle adjustment.

\subsection{Depth-Guided Global Optimization}
After view-graph and correspondence filtering, the remaining image pairs provide reliable scale-aware relative poses and depth-consistent feature tracks.
We use these constraints to initialize the global reconstruction.

\PAR{Focal Averaging.}
The depth-aware relative pose solver provides pairwise intrinsics estimates $(\mathbf{K}_i, \mathbf{K}_j)$ (Eqn.\ref{eqn:solver}) for the matched image pairs.
Pixels are assumed to be squared and principal points  coincident with image centers~\cite{hartley2012efficient, dust3r}, leaving only focal lengths to estimate. 
Due to image noise and inaccuracies in the monocular depth estimates, each edge connected to camera $I_i$ yields a different estimate of that camera's focal length. 
We therefore aggregate the pairwise focal estimates into a single robust focal length for each image.

For each verified edge $(i,j)\in\mathcal{E}$, let $\hat{f}_{i}^{ij}$ and $\hat{f}_{j}^{ij}$ denote the focal lengths of images $I_i$ and $I_j$ estimated from the depth-aware relative pose solver. 
For each image $I_i$, we collect all focal estimates from its incident edges:
\begin{equation}
\mathcal{F}_i
=
\{
\hat{f}_{i}^{ij}\mid (i,j)\in\mathcal{E}
\}
\cup
\{
\hat{f}_{i}^{ji}\mid (j,i)\in\mathcal{E}
\} \enspace.
\end{equation}
We then compute the final focal length, which serves as a stable initialization for subsequent global optimization and bundle adjustment using a median estimator:
$f_i = \operatorname{median}\left(\mathcal{F}_i\right) \enspace.$

\PAR{Global Scale Averaging.}
From the depth-aware relative pose solver~\cite{ding2025reposed} of Eqn.~\ref{eqn:solver}, each retained edge provides a scaled relative pose $(\mathbf{R}_{ij}, \mathbf{t}_{ij})$, and also a pairwise depth-scale ratio $s_{ij}$.
Since monocular depth predictions may still contain image-wise scale inconsistencies, we recover globally consistent scale factors over the filtered view graph. 
For each retained edge $(i, j)$, the pairwise scale ratio provides the constraint $s_{ij} = {s_j}/{s_i}$.
Taking the logarithm gives a linear relation in log-scale space:
\begin{equation}
\log s_{ij} = \log s_j - \log s_i \enspace.
\end{equation}
For simplicity, we denote $\ell_i=\log s_i$. 
The global image-wise scales are then estimated by solving the robust scale-averaging problem
\begin{equation}
{\ell_i^\star}
=
\arg\min_{\ell_i}
\sum_{(i,j)\in\mathcal{E}}
\rho_s
\left(
\ell_j-\ell_i-\ell_{ij}
\right) \enspace,
\end{equation}
where $\rho_s(\cdot)$ is a robust loss.
The scale gauge is fixed by setting $\ell_1 = 0$, or equivalently $s_1 = 1$. 
The final global depth scale of each image is obtained as
$s_i = \exp(\ell_i^\star)$.
The estimated scales are then used to align all monocular depth maps into a common reconstruction scale,
$\tilde{D}_i(\mathbf{x}) = s_i D_i(\mathbf{x})$.
These scale-consistent depth maps provide metric geometric priors for subsequent camera positions and 3D points initialization.

\PAR{Depth-Scaled Pose-Point Initialization.}
After global scale averaging, we use the scaled relative poses and scaled monocular depths to initialize the cameras and 3D points before GLOMAP~\cite{glomap} global positioning.
We construct a maximum spanning tree (MST)~\cite{kruskal1956shortest} from the filtered view graph and initialize camera centers by chaining scaled relative translations along the tree.
Let $\mathcal{M}\subset\mathcal{E}$ denote the MST, where edges are weighted by their number of inliers. 
We select a root image $I_r$ and fix its camera center as 
$\mathbf{c}_r^{0}=\mathbf{0}$. 
For each tree edge $(i,j)\in\mathcal{M}$, we convert the pairwise scaled relative translation $\mathbf{t}_{ij}$ into the common reconstruction scale using the global image scale $s_i$:
\begin{equation}
\tilde{\mathbf{t}}_{ij}=s_i\mathbf{t}_{ij} \enspace.
\end{equation}
Given global rotations ${\mathbf{R}_i}$ from rotation averaging, the relation between camera centers and the relative translation is 
$\tilde{\mathbf{t}}_{ij}
=
\mathbf{R}_j(\mathbf{c}_i-\mathbf{c}_j)$. 
Therefore, if image $i$ is the parent of image $j$ in the MST, the child camera center is initialized as
\begin{equation}
\mathbf{c}_j^{0}
=
\mathbf{c}_i^{0}
-
\mathbf{R}_j^{\top}\tilde{\mathbf{t}}_{ij} \enspace.
\end{equation}
By recursively applying this update from the root to all nodes in the MST, we obtain initial camera centers $\{\mathbf{c}_i^{0}\}_{i=1}^{N}$.
We then initialize sparse 3D points by lifting feature observations using the globally scaled monocular depth maps.
For a track $p$ observed in image $i$, the 2D feature $\mathbf{x}_{(i,p)}$ with its homogeneous coordinate $\bar{\mathbf{x}}_{(i,p)}$ can be lifted with the globally averaged focal $f_i$ as follows:
\begin{equation}
\mathbf{u}_{(i, p)}
=
\frac{\mathbf{K}_i^{-1}\bar{\mathbf{x}}_{(i, p)}}
{\left\|\mathbf{K}_i^{-1}\bar{\mathbf{x}}_{(i, p)}\right\|_2} \enspace.
\end{equation}
This observation is transformed into the world coordinate system as
\begin{equation}
\mathbf{X}_{(i, p)}^{w}
=
\mathbf{c}_i^{0}
+
\mathbf{R}_i^{\top}
\left(
\tilde{D}_i(\mathbf{x}_{(i, p)})\mathbf{u}_{(i, p)}
\right) \enspace.
\end{equation}
For a track $p$ observed in multiple images, we initialize the 3D point by aggregating its lifted observations: 
\begin{equation}
\mathbf{X}_p^{0}
=
\operatorname{median}
\left(
\{
\mathbf{X}_{(i, p)}^{w}
\mid
i\in\mathcal{I}(p)
\}
\right) \enspace,
\end{equation}
where $\mathcal{I}(p)$ is the set of images observing track $p$. 

The initialized camera centers ${\mathbf{c}_i^{0}}$ and depth-lifted points ${\mathbf{X}_p^{0}}$ are then passed to the standard GLOMAP global positioning stage, which jointly refines camera centers and 3D points using multi-view ray consistency.:
\begin{equation}
\arg\min_{{\mathbf{c}_i},{\mathbf{X}_p}}
\sum_{(i,p)\in\mathcal{O}}
\rho
\left(
\alpha_{(i, p)}(\mathbf{X}_p-\mathbf{c}_i)
,
\mathbf{R}_i^\top\mathbf{u}_{(i, p)}
\right) \enspace,
\end{equation}
where $\mathcal{O}$ is the set of feature observations, $\alpha$ is a normalizing factor, and $\rho$ is a robust loss.
The final reconstruction is still refined by global bundle adjustment.

\section{Experiments}
\label{sec:experiments}

\PAR{Implementation Details.}
We implement DGSfM in Python using Numpy and PyTorch with GPU-accelerated sparse optimization based on BAE~\cite{zhan2026bundle}.
MoGe2~\cite{wang2025moge2} is used to extract metric depth maps for the main experiments.
We run scale averaging, global positioning, and bundle adjustment for 30, 100, and 100 iterations, respectively.
Bundle adjustment is repeated three times to further refine camera parameters and 3D structure.
More 
details, including error thresholds and filtering parameters are  in the appendix. 

\PAR{Sparse Detectors and Matching.}
We experiment with several sparse feature detection and matching methods, including SIFT~\cite{sift} with nearest-neighbor matching, ALIKED~\cite{zhao2023aliked} with LightGlue~\cite{lindenberger2023lightglue} matching, and the recent LoMa~\cite{nordstrom2026loma} detection and matching.
The maximum number of extracted features is set to 8192, 2048, and 4096 for SIFT, ALIKED, and LoMa, respectively.

\PAR{Dense Matchers.}
We also experiment with dense matching methods, including RoMa v1~\cite{romav1} and RoMa v2~\cite{romav2}.
For each image pair, we sample 10,000 correspondences from the dense matches.
Unlike previous dense feature refinement methods~\cite{dfsfm, densesfm}, we do not apply match quantization~\cite{dfsfm} or non-maximum suppression~\cite{densesfm, superpoint} which shifts the matches when selecting dense correspondences.
Instead, we rely on our proposed filtering strategy to remove unreliable matches before track construction and reconstruction.

\begin{table*}[t]\centering
\caption{
Multi-View Camera Pose Estimation on ETH-3D~\cite{eth3dsfm} and IMC-2021~\cite{imc2021}. We compare DGSfM (Ours) with traditional SfM pipelines (incremental (I) and global (G)), feature refinement and dense-matching-based methods (R), monocular depth map-based approaches (D), and feed-forward 3D reconstruction approaches (F) across different feature and matching front-ends. Ours$^\dagger$ denotes our method without the Doppelgangers++~\cite{doppelgangers++} module. \colorbox{pastel_yellow}{Best}, \colorbox{pastel_blue}{Second}, and \colorbox{pastel_pink}{Third} results are highlighted.	
}
\label{tab:eth3dimc}
\resizebox{\textwidth}{!}{
\begin{tabular}{l|c|c|c|*{3}{c}|*{3}{c}}
\toprule
\multirow{2}{*}{Method} & \multirow{2}{*}{Type} & \multirow{2}{*}{Feature} & \multirow{2}{*}{Matching} & \multicolumn{3}{c|}{ETH-3D~\cite{eth3dsfm}} & \multicolumn{3}{c}{IMC-2021~\cite{imc2021}} \\ 
& & & & AUC@$1\degree$ & AUC@$3\degree$ & AUC@$5\degree$ & AUC@$3\degree$ & AUC@$5\degree$ & AUC@$10\degree$ \\
\midrule
COLMAP~\cite{colmap} & I & \multirow{4}{*}{SIFT~\cite{sift}} & \multirow{4}{*}{NN} & \Third 31.58 & \Third 45.65 & \Third 49.18 & \Third 27.56 & \Third 37.10 & \Third 48.55 \\
GLOMAP~\cite{glomap} & G  & & & \Best 41.22 & \Second 56.56 & \Second 60.58 & \Second 33.79 & \Second 44.48 & \Second 57.38 \\
InstantSfM~\cite{instantsfm} & G  & & & 18.95 & 33.27 & 39.11 & - & - & - \\
\textbf{Ours}$^\dagger$ & D+G  & & & \Second 38.88 & \Best 61.67 & \Best 70.12 & \Best 35.55 & \Best 48.29 & \Best 64.87 \\
\midrule
COLMAP~\cite{colmap} & I  & \multirow{3}{*}{ALIKED~\cite{zhao2023aliked}} & \multirow{3}{*}{LightGlue~\cite{lindenberger2023lightglue}} & \Third 35.50 & \Second 53.77 & \Second 58.68 & \Third 38.89 & \Third 50.83 & \Third 64.40 \\
GLOMAP~\cite{glomap} & G  & & & \Second 36.23 & \Third 50.97 & \Third 55.21 & \Second 41.31 & \Second 53.56 & \Second 67.37 \\
\textbf{Ours}$^\dagger$ & D+G  & & & \Best 37.19 & \Best 59.43 & \Best 66.96 & \Best 43.43 & \Best 56.45 & \Best 71.54 \\
\midrule
COLMAP~\cite{colmap} & I  & \multirow{5}{*}{LoMa-B~\cite{nordstrom2026loma}} & \multirow{5}{*}{LoMa-B~\cite{nordstrom2026loma}} & 43.45 & \Third 61.66 & \Third 66.32 & 42.40 & 54.27 & 67.36 \\
GLOMAP~\cite{glomap} & G  & & & \Third 43.74 & 60.19 & 64.84 & \Third 45.91 & \Third 58.29 & \Third 71.81 \\
InstantSfM~\cite{instantsfm} & G  & & & 23.44 & 34.64 & 36.78 & - & - & - \\
\textbf{Ours}$^\dagger$ & D+G  & & & \Second 46.87 & \Second 71.92 & \Second 78.78 & \Best 46.62 & \Best 60.84 & \Second 73.80 \\
\textbf{Ours} & D+G  & & & \Best 51.60 & \Best 78.25 & \Best 85.43 & \Second 46.59 & \Second 59.91 & \Best 75.59 \\
\midrule
COLMAP~\cite{colmap} & I & \multirow{5}{*}{LoMa-G~\cite{nordstrom2026loma}} & \multirow{5}{*}{LoMa-G~\cite{nordstrom2026loma}} & \Third 48.19 & \Third 65.61 & \Third 71.12 & \Third 44.58 & 56.18 & 68.73 \\
GLOMAP~\cite{glomap} & G  & & & 46.01 & 61.70 & 66.37 & \Second 46.75 & \Third 59.32 & \Third 72.36 \\
InstantSfM~\cite{instantsfm} & G  & & & 29.72 & 38.83 & 42.75 & - & - & - \\
\textbf{Ours}$^\dagger$ & D+G  & & & \Second 49.01 & \Second 72.97 & \Second 79.31 & \Second 46.75 & \Second 60.16 & \Second 74.75 \\
\textbf{Ours} & D+G  & & & \Best 53.58 & \Best 79.11 & \Best 86.04 & \Best 47.74 & \Best 61.12 & \Best 75.65 \\
\midrule \midrule
MP-SfM~\cite{mpsfm} & D+I  & SuperPoint~\cite{superpoint} & RoMa v1~\cite{romav1} & 5.64 & 22.44 & 34.16 & 19.43 & 29.13 & 40.87 \\
PixSfM~\cite{pixsfm} & I+R  & \multicolumn{2}{|c|}{feature metric refinement  of SIFT~\cite{sift}} & 26.94 & 39.01 & 42.19 & 25.54 & 34.80 & 46.73 \\ 
MASt3R-SfM~\cite{mast3rsfm} & D+G & MASt3R~\cite{mast3r} & MASt3R~\cite{mast3r} & 27.05 & 45.84 & 52.61 & 31.77 & 46.36 & 64.37 \\ \cline{3-4}
DF-SfM~\cite{dfsfm} & I+R  & \multirow{4}{*}{RoMa v1~\cite{romav1}} & \multirow{4}{*}{RoMa v1~\cite{romav1}} & \Third 59.12 & \Third 75.59 & \Third 79.53 & 47.43 & 59.84 & 73.19 \\
Dense-SfM~\cite{densesfm} & I+R  & & & \Second 60.92 & \Second 78.41 & \Second 82.63 & \Third 48.48 & \Third 60.79 & \Third 73.90 \\
MV-RoMa~\cite{mvroma} & I+R & & & - & - & - & \Best 51.31 & \Second 62.92 & \Second 75.92 \\
\textbf{Ours} & D+G & & & \Best 61.00 & \Best 82.63 & \Best 88.33 & \Second 50.66 & \Best 63.03 & \Best 76.30 \\
\midrule \midrule
VGGT~\cite{vggt} & F & \multirow{3}{*}{-} & \multirow{3}{*}{-} & 0.65 & 8.37 & 17.76 & 41.15 & 55.52 & 71.73 \\
Pi3~\cite{pi3} & F & & & 14.31 & 38.63 & 51.25 & 43.34 & 57.57 & 73.35 \\
DepthAnything3~\cite{lin2025depthanything3} & F & & & 15.53 & \Third 46.27 & \Third 58.63 & 44.15 & 58.54 & \Third 74.21 \\ \cline{3-4}
VGGT~\cite{vggt}+BA & F & ALIKED~\cite{zhao2023aliked}+SP~\cite{superpoint} & VGGSfM~\cite{vggsfm} & 18.23 & 38.49 & 46.98 & 44.96 & 58.08 & 73.52 \\
Pi3~\cite{pi3}+BA & F & SIFT~\cite{sift}+Deep & VGGSfM~\cite{vggsfm}+Deep & \Second 32.84 & \Second 58.51 & \Second 68.19 & \Best 55.30 & \Best 66.43 & \Best 78.67 \\
VGGSfM~\cite{vggsfm} & F+G & SP~\cite{superpoint}+SIFT~\cite{sift} & Deep & \Third 28.09 & 44.41 & 49.91 & \Third 45.23 & \Third 58.89 & 73.92 \\ 
\textbf{Ours} (Best) & D+G & RoMa v1~\cite{romav1} & RoMa v1~\cite{romav1} & \Best 61.00 & \Best 82.63 & \Best 88.33 & \Second 50.66 & \Second 63.03 & \Second 76.30 \\
\bottomrule
\end{tabular}
}
\vspace{-6mm}
\end{table*}

\begin{figure*}[t]
  \centering
\includegraphics[width=\textwidth]{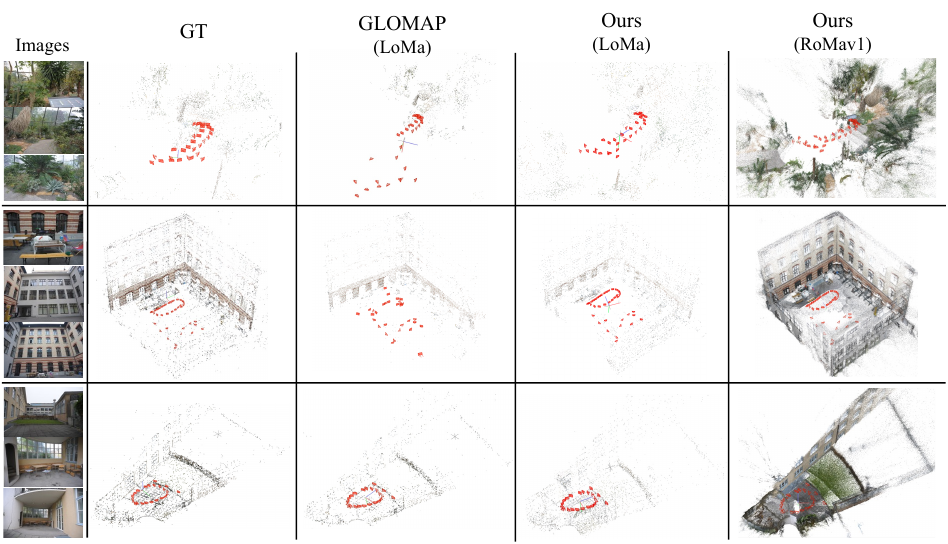}
  \caption{Qualitative comparison on ETH3D~\cite{eth3dsfm}. Compared with GLOMAP~\cite{glomap}, our method produces more stable camera pose estimates with LoMa~\cite{nordstrom2026loma} matches, while using dense RoMa v1~\cite{romav1} matches leads to more complete 3D structure.
  }
  \label{fig:eth3d_results}
  \vspace{-4mm}
\end{figure*}

\subsection{Multi-view Camera Pose Estimation}
We compare our method with three groups of baselines.
1) Traditional SfM pipelines~\cite{colmap, glomap, instantsfm}.
2) Feature refinement~\cite{pixsfm} and dense-matching-based SfM methods~\cite{dfsfm, densesfm, mvroma}.
3) Feed-forward 3D reconstruction methods~\cite{vggsfm, vggt, pi3, lin2025depthanything3, mast3rsfm}. 
We also compare with MP-SfM~\cite{mpsfm}, that uses monocular priors but requires intrinsics as input.
For all methods, we assume that camera intrinsics are not provided as input. 
Since MP-SfM assumes known camera intrinsics, we instead provide it with focal lengths estimated by the mono-depth model, MoGe2~\cite{wang2025moge2} (also used to get monocular depth and surface normal priors) to enable a fair comparison with other methods.
We evaluate camera pose accuracy using the area under the curve (AUC) of the pose error~\cite{imc2021} under multiple thresholds depending on the dataset, following standard practice~\cite{pixsfm, dfsfm, densesfm}.

\PAR{ETH3D and IMC2021.}
Tab.~\ref{tab:eth3dimc} reports results on the ETH3D~\cite{eth3dsfm} and IMC2021 Phototourism~\cite{imc2021} datasets. 
ETH3D contains indoor and outdoor scenes captured with sparse high-resolution images, with accurate LiDAR-aligned camera poses provided as ground truth. 
IMC2021 consists of large-scale outdoor scenes and is evaluated using image bags of different sizes, including 5, 10, and 25 images, rather than the full image collection.
Evaluation of both datasets follows the same protocol as in ~\cite{dfsfm, densesfm}.
For a fair comparison with traditional SfM methods, rows marked by Ours$^\dagger$ disable the Doppelgangers++~\cite{doppelgangers++} module. 

Compared with traditional SfM pipelines, our method achieves comparable or better performance at the strictest threshold (AUC@$1\degree$) while providing substantial improvements at the more relaxed AUC thresholds (AUC@$3\degree$ and AUC@$5\degree$), 
as further illustrated by the qualitative results in Fig.~\ref{fig:eth3d_results}.
Among the sparse feature extraction and matching methods, our method with LoMa~\cite{nordstrom2026loma} matches achieves the best performance, on par with dense feature refinement methods: DF-SfM~\cite{dfsfm} and Dense-SfM~\cite{densesfm}.
With dense matches, our method also obtains comparable accuracy compared to dense feature refinement methods on all thresholds without having additional match refinement architectures.
Feed-forward reconstruction methods perform poorly on ETH3D, but achieve competitive results on IMC2021 compared with sparse feature-based SfM methods, due to small image collections with high overlap.

\PAR{Runtimes.}
Tab.~\ref{tab:ablation_runtime} compares runtimes on the ETH3D dataset. Our method has a runtime comparable to GLOMAP while being substantially faster than MP-SfM. For methods that rely on monocular depth estimation, the reported runtime is split into the SfM runtime and the processing time required to obtain monocular depth estimates for all images. 
The latter takes 3.35s on ETH3D.

\subsection{Ablation Study}
Tables~\ref{tab:ablation_depth} and~\ref{tab:ablation} present ablation studies of our pipeline using LoMa-B~\cite{nordstrom2026loma} matches on the ETH3D~\cite{eth3dsfm} dataset.

\begin{table}[t]
\centering

\begin{minipage}[t]{0.42\linewidth}
\centering
\caption{Runtime comparison.}
\label{tab:ablation_runtime}
\scriptsize{
\begin{tabular}{l|c}
\toprule
Method & Time (s) \\
\midrule
COLMAP~\cite{colmap} & 12.70 \\
GLOMAP~\cite{glomap} & 16.40 \\
MP-SfM~\cite{mpsfm} & 110.58+3.35 \\
Ours & 13.99+3.35 \\
\bottomrule
\end{tabular}
}%
\end{minipage}
\hfill
\begin{minipage}[t]{0.54\linewidth}
\centering
\caption{Ablation with different depth models.}
\label{tab:ablation_depth}
\scriptsize{
\begin{tabular}{l|ccc}
\toprule
Method & AUC@$1\degree$ & AUC@$3\degree$ & AUC@$5\degree$ \\
\midrule
MoGe2~\cite{wang2025moge2} & 51.60 & 78.25 & 85.43 \\
MoGe1~\cite{wang2025moge} & 52.47 & 78.34 & 85.53 \\
UniK3D~\cite{piccinelli2025unik3d} & 49.00 & 75.70 & 83.23 \\
UniDepthV2~\cite{piccinelli2025unidepthv2} & 48.17 & 73.99 & 81.29 \\
\bottomrule
\end{tabular}
}%
\end{minipage}

\end{table}



\begin{wraptable}[9]{r}{0.52\linewidth}
\centering
\vspace{-14mm}
\caption{Ablation study of main modules.}
\vspace{1mm}
\label{tab:ablation}
\scriptsize{ 
\begin{tabular}{l|*{3}{c}}
\toprule
Method & AUC@$1\degree$ & AUC@$3\degree$ & AUC@$5\degree$ \\
\midrule
Full & 51.60 & 78.25 & 85.43 \\
\midrule
w/o doppelgangers++ & 46.87 & 71.92 & 78.78 \\
w/o triplet-edge prune & 49.54 & 76.22 & 83.41 \\
w/o scaled-depth filter & 48.61 & 74.46 & 81.73 \\
\midrule
w/o global positioning & 40.70 & 63.40 & 70.64 \\
w/o scale averaging & 47.54 & 72.98 & 80.16 \\
w/o depth pose init. & 48.81 & 74.03 & 81.18 \\[-0.8ex]
 & {\tiny $\quad \pm$(0.84)} & {\tiny $\quad \pm$(1.04)} & {\tiny $\quad \pm$(0.97)} \\
\bottomrule
\end{tabular}
}%
\end{wraptable}

\PAR{View-Graph Filtering.}
As shown in Tab.~\ref{tab:eth3dimc} and Tab.~\ref{tab:ablation}, removing the Doppelgangers++~\cite{doppelgangers++} stage causes a clear performance drop, especially at relaxed thresholds on ETH3D, showing that visually ambiguous image pairs can still pass geometric verification and negatively affect global reconstruction.
Triplet-guided edge pruning also improves performance, confirming that local three-view consistency helps remove unreliable view-graph edges that are difficult to detect from pairwise checks alone.
Without the scaled-depth consistency filter, the AUC also decreases, indicating that epipolar verification alone is insufficient to remove all incorrect matches.

\PAR{Global Optimization.}
Tab.~\ref{tab:ablation} also ablates the global positioning stage by directly applying bundle adjustment (BA) after depth-scaled pose-point initialization.
This leads to a large performance drop, indicating that the initialization alone is insufficient and that global positioning is essential for jointly refining camera centers and 3D points using multi-view ray consistency before a final BA.
Global scale averaging is also critical among the depth-related components.
Removing it reduces the AUC@5 from $85.43$ to $80.16$, showing that pairwise depth-scale estimates must be globally aligned before they can provide reliable depth lifting and initialization.
Replacing our depth-scaled pose-point initialization with random initialization also degrades performance and, importantly, introduces run-to-run variation, as shown by the reported standard deviations.

\PAR{Sensitivity to Monocular Depth Quality.} 
Tab.~\ref{tab:ablation_depth} shows results obtained by using different monocular depth estimators inside our approach. 
As can be expected, depth prediction accuracy affects SfM performance, with MoGe1 
and MoGe2 
leading to the best results. 
Still, even the "worst" predictor (UniDepthV2) 
improves performance compared to COLMAP, GLOMAP, and InstantSfM (as evident from comparing Tab.~\ref{tab:ablation_depth} with results obtained with LoMa-B in Tab.~\ref{tab:eth3dimc}).  




\section{Conclusion}
This paper presented DGSfM, a depth-guided global SfM pipeline that incorporates monocular geometry into global reconstruction while retaining explicit multi-view optimization.
By estimating scale-aware relative poses, filtering view-graph edges and correspondences with depth consistency, averaging image-wise depth scales, and initializing cameras and 3D points from scaled monocular depths, DGSfM reduces the scale ambiguity and initialization sensitivity of conventional global SfM. 
Extensive experiments demonstrate consistent gains over strong SfM baselines across both sparse and dense matching front-ends, indicating that monocular geometry can provide useful metric constraints while preserving the accuracy and scalability of explicit geometric optimization.

\section{Limitations and Future Work}
Similarly to other SfM methods that use depth maps~\cite{mast3rsfm,mpsfm}, our reliance on monocular depth estimators introduces a potential error source (\eg, inaccurate or wrong depth estimates). 
In practice, this does not seem to be a critical limitation, as monocular depth estimators are constantly improving. 
The issue can be further alleviated by taking depth prediction uncertainty into account during relative pose estimation and filtering, which we leave for future work. 
In this context, jointly optimizing over 3D-2D reprojection errors and 2D-2D epipolar errors could also help handle inaccuracies in the depth maps. 
Our approach currently uses the depth maps to improve the SfM process. 
Thus, an interesting direction for future work is to use the results of the SfM process to improve the depth maps to obtain more accurate dense reconstructions. 

\newpage
\clearpage
\appendix
\begin{center}
    \begingroup
    \centering
    {\LARGE \bf Appendix \par}
    \endgroup
\end{center}

\setcounter{section}{0}
\setcounter{equation}{0}
\setcounter{figure}{0}
\setcounter{table}{0}

\renewcommand{\thesection}{A\arabic{section}}
\renewcommand{\theequation}{A\arabic{equation}}
\renewcommand{\thefigure}{A\arabic{figure}}
\renewcommand{\thetable}{A\arabic{table}}

In this appendix, we provide additional details and results to support the main paper.
We first describe the experimental setup of our method and the baselines used for fair comparison in Sec. \ref{sec:exp_details}. 
We then present additional experiments in Sec. \ref{sec:add_exp}.
Finally, we provide further qualitative comparisons with existing methods in Sec. \ref{sec:add_vis}.

\section{Experimental Setup}
\label{sec:exp_details}

\PAR{Hardware Details.}
The experiments are run on a server with Intel Xeon Gold 6150 and a Nvidia-V100-32GB GPU.

\PAR{Hyperparameters.}
For depth-aware relative pose estimation with RePoseD~\cite{ding2025reposed}, we set the Sampson error threshold to $1.0$, the reprojection error threshold to $16.0$, and the number of RANSAC iterations to $1000$.
For pairwise edge pruning with Dopplegangers++~\cite{doppelgangers++}, we use a confidence threshold of $\tau_{DG} = 0.8$.
For triplet-guided edge pruning, we set the triplet support threshold to $\tau_{tri} = 0.6$.
In the scaled depth consistency filtering stage, the reprojection error threshold $\tau_{\pi}$ and depth consistency threshold $\tau_d$ are set to $12.0$ and $0.3$, respectively.
We further require each retained image pair to have at least $15$ inliers and an inlier ratio of at least $0.25$.
For track construction, the minimum number of views required for a track is set to 3 views for sparse feature detectors and 2 views for dense matchers.

\PAR{Dataset and Evaluation Details.}
We follow DF-SfM~\cite{dfsfm} and Dense-SfM~\cite{densesfm} to evaluate on both ETH3D~\cite{eth3dsfm} and IMC-2021~\cite{imc2021} Phototourism benchmarks.
For ETH3D, we use 22 indoor and outdoor scenes with images resized to $1600\times 1056$.
For IMC2021, there are 9 scenes in total but evaluation is performed on overlapping image subsets rather than the full image collections.
Specifically, each scene is evaluated using 5-image, 10-image, and 25-image bags, with 100, 50, and 25 different subset variations, respectively.
As a result, the overall performance is strongly influenced by the 5-image bags, which tend to favor feed-forward reconstruction methods with multi-view supervision that are effective on small local view graphs.
For all scenes, we use exhaustive matching to construct the initial image pairs.

\section{Experimental Evaluation}
\label{sec:add_exp}

\begin{table*}[t]\centering
\caption{
Multi-view Camera Pose Estimation on LaMAR~\cite{sarlin2022lamar}. \colorbox{pastel_yellow}{Best} and \colorbox{pastel_blue}{Second-best} results are highlighted.
}
\label{tab:lamar}
\resizebox{\textwidth}{!}{
\begin{tabular}{l|c|c|c|*{3}{c}|*{3}{c}|*{3}{c}}
\toprule
\multirow{2}{*}{Method} & \multirow{2}{*}{Type} & \multirow{2}{*}{Feature} & \multirow{2}{*}{Matching} & \multicolumn{3}{c|}{LIN} & \multicolumn{3}{c}{HGE} \\ 
& & & & AUC@$5\degree$ & AUC@$10\degree$ & AUC@$30\degree$ & AUC@$5\degree$ & AUC@$10\degree$ & AUC@$30\degree$ \\
\midrule
COLMAP~\cite{colmap} & I & \multirow{3}{*}{SIFT~\cite{sift}} & \multirow{3}{*}{NN} & \Second 10.69 & \Second 11.62 & \Second 12.66 & \Second 5.34 & \Second 5.65 & \Second 5.92 \\
GLOMAP~\cite{glomap} & G & & & 5.11 & 7.19 & 11.09 & 1.33 & 1.71 & 2.96 \\
\textbf{Ours} & D+G & & & \Best 14.48 & \Best \Best 24.78 & \Best 31.42 & \Best 9.32 & \Best 12.43 & \Best 20.64 \\
\midrule
COLMAP~\cite{colmap} & I & \multirow{3}{*}{LoMa-B~\cite{nordstrom2026loma}} & \multirow{3}{*}{LoMa-B~\cite{nordstrom2026loma}} & \Second 24.76 & \Second 32.81 & \Second 49.07 & \Second 4.80 & \Second 5.43 & \Second 7.30 \\
GLOMAP~\cite{glomap} & G & & & 21.18 & 29.93 & 43.65 & 1.07 & 1.51 & 3.49 \\
\textbf{Ours} & D+G & & & \Best 32.45 & \Best 45.58 & \Best 63.74 & \Best 10.49 & \Best 15.37 & \Best 23.49 \\

\bottomrule
\end{tabular}
}
\end{table*}

\PAR{Results on LaMAR.}
We further evaluate our method on the large-scale LaMAR~\cite{sarlin2022lamar} dataset.
We report results on two scenes: LIN, an outdoor scene, and HGE, a mixed indoor-outdoor scene.
Each scene contains tens of thousands of images captured by different AR devices and smartphones, resulting in large-scale image collections with diverse viewpoints and challenging appearance variations.
From each scene, we select the front-facing images captured by the NavVis device for evaluation, resulting in 993 images for LIN and 1,072 images for HGE.
Tab.~\ref{tab:lamar} reports the results.
For SIFT-NN matching, we use the default \texttt{vocab\_tree\_matcher} in COLMAPv4.0 to generate image pairs.
For LoMa matching, we use MegaLoc~\cite{Berton_2025_MegaLoc} retrieval and keep the top-50 retrieved image pairs for each image.

Our method consistently outperforms COLMAP and GLOMAP on both scenes and with both matching front-ends.
The gains are particularly strong on LIN, where the outdoor scene provides more stable visual overlap and long-range connectivity.
Overall performance on HGE is lower for all methods, reflecting the higher difficulty of this mixed indoor-outdoor scene, where appearance changes and weaker global connectivity make reliable image retrieval and view-graph construction more challenging.
Nevertheless, our method still provides clear improvements.

\begin{table*}[t]\centering
\caption{
Multi-View Camera Pose Estimation on ETH-3D~\cite{eth3dsfm} and IMC-2021~\cite{imc2021} with estimated and ground-truth intrinsics. Ours uses intrinsics from focal averaging, and MP-SfM uses MoGe2~\cite{wang2025moge2} estimated intrinsics as input. Ours$^\star$ and MP-SfM$^\star$ use ground-truth intrinsics. \colorbox{pastel_yellow}{Best} and \colorbox{pastel_blue}{Second-best} results are highlighted.
}
\label{tab:eth3dimc_calib}
\resizebox{\textwidth}{!}{
\begin{tabular}{l|c|c|c|*{3}{c}|*{3}{c}}
\toprule
\multirow{2}{*}{Method} & \multirow{2}{*}{Type} & \multirow{2}{*}{Feature} & \multirow{2}{*}{Matching} & \multicolumn{3}{c|}{ETH-3D~\cite{eth3dsfm}} & \multicolumn{3}{c}{IMC-2021~\cite{imc2021}} \\ 
& & & & AUC@$1\degree$ & AUC@$3\degree$ & AUC@$5\degree$ & AUC@$3\degree$ & AUC@$5\degree$ & AUC@$10\degree$ \\
\midrule
MP-SfM~\cite{mpsfm} & D+I  & SuperPoint~\cite{superpoint} & RoMa v1~\cite{romav1} & 5.64 & 22.44 & 34.16 & 19.43 & 29.13 & 40.87 \\
\textbf{Ours} & D+G  & LoMa-B~\cite{nordstrom2026loma} & LoMa-B~\cite{nordstrom2026loma} & \Second 51.60 & \Second 78.25 & \Second 85.43 & \Second 46.59 & \Second 59.91 & \Second 75.59 \\
\textbf{Ours} & D+G & RoMa v1~\cite{romav1} & RoMa v1~\cite{romav1} & \Best 61.00 & \Best 82.63 & \Best 88.33 & \Best 50.66 & \Best 63.03 & \Best 76.30 \\
\midrule
MP-SfM$^\star$~\cite{mpsfm} & D+I  & SuperPoint~\cite{superpoint} & RoMa v1~\cite{romav1} & \Second 58.63 & 78.87 & 84.34 & \Best 56.27 & 62.78 & 70.54 \\
\textbf{Ours}$^\star$ & D+G  & LoMa-B~\cite{nordstrom2026loma} & LoMa-B~\cite{nordstrom2026loma} & 56.74 & \Second 81.16 & \Second 87.71 & 50.77 & \Second 63.85 & \Second 77.68 \\
\textbf{Ours}$^\star$ & D+G & RoMa v1~\cite{romav1} & RoMa v1~\cite{romav1} & \Best 64.69 & \Best 85.20 & \Best 90.50 & \Second 55.28 & \Best 68.71 & \Best 82.36 \\
\bottomrule
\end{tabular}
}
\end{table*}

\PAR{Using Ground-Truth Intrinsics.}
Tab.~\ref{tab:eth3dimc_calib} highlights the importance of camera calibration for monocular-prior-based SfM.
When ground-truth intrinsics are not provided, MP-SfM~\cite{mpsfm} shows limited performance, especially on ETH3D, indicating that its incremental reconstruction pipeline is highly sensitive to accurate calibration.
In contrast, our method remains robust with intrinsics estimated from the depth-aware relative pose solver~\cite{ding2025reposed} and focal averaging, achieving strong performance on both ETH3D and IMC2021.

When ground-truth intrinsics are available, MP-SfM improves substantially, confirming that accurate calibration is critical for its two-view initialization, next-view registration, and bundle adjustment.
Our method also benefits from ground-truth intrinsics and achieves the best overall performance with both LoMa and RoMa matches.
Our method outperforms MP-SfM at almost all thresholds with dense RoMa matches.
This shows that better calibration further strengthens depth-guided global optimization.

\begin{table*}[t]\centering
\caption{
Point Cloud Accuracy on ETH-3D~\cite{eth3dsfm}. We compare DGSfM (Ours) with traditional SfM pipelines (incremental (I) and global (G)), feature refinement and dense-matching-based methods (R), monocular depth map-based approaches (D), and feed-forward 3D reconstruction approaches (F) across different feature and matching front-ends. \colorbox{pastel_yellow}{Best}, \colorbox{pastel_blue}{Second}, and \colorbox{pastel_pink}{Third} results are highlighted.	
}
\label{tab:eth3d_triang}
\resizebox{\textwidth}{!}{
\begin{tabular}{l|c|c|c|*{3}{c}|*{3}{c}}
\toprule
\multirow{2}{*}{Method} & \multirow{2}{*}{Type} & \multirow{2}{*}{Feature} & \multirow{2}{*}{Matching} & \multicolumn{3}{c|}{Accuracy (\%)} & \multicolumn{3}{c}{Completeness (\%)} \\ 
& & & & \SI{1}{\centi\metre} & \SI{2}{\centi\metre} & \SI{5}{\centi\metre} & \SI{1}{\centi\metre} & \SI{2}{\centi\metre} & \SI{5}{\centi\metre} \\
\midrule
COLMAP~\cite{colmap} & I & \multirow{3}{*}{LoMa-B~\cite{nordstrom2026loma}} & \multirow{3}{*}{LoMa-B~\cite{nordstrom2026loma}} & \Second 31.84 & \Third 43.21 & \Third 58.66 & \Best 0.66 & \Best 2.93 & \Best 13.28 \\
GLOMAP~\cite{glomap} & G  & & & \Third 31.65 & \Second 45.82 & \Second 65.07 & \Second 0.27 & \Second 1.50 & \Second 9.14 \\
\textbf{Ours} & D+G  & & & \Best 45.28 & \Best 61.25 & \Best 79.76 & \Third 0.26 & \Third 1.32 & \Third 7.49 \\
\midrule \midrule
DepthAnything3~\cite{lin2025depthanything3} & F &  &  & 13.62 & 23.96 & 44.44 & \Third 8.08 & 16.24 & 27.12 \\ \cline{3-4}
PixSfM~\cite{pixsfm} & I+R  & \multicolumn{2}{c|}{featuremetric refinement  of SIFT~\cite{sift}} & 76.18 & 85.60 & 93.16 & 0.17 & 0.71 & 3.29 \\
VGGSfM~\cite{vggsfm} & F+G & SP~\cite{superpoint}+SIFT~\cite{sift} & Deep & \Third 80.62 & \Third 89.49 & \Third 96.52 & 4.52 & 13.11 & 33.96 \\
MASt3R-SfM~\cite{mast3rsfm} & D+G & MASt3R~\cite{mast3r} & MASt3R~\cite{mast3r} & 27.34 & 43.90 & 69.50 & \Second 14.86 & \Second 34.81 & \Second 68.72 \\ 
MP-SfM~\cite{mpsfm} & D+I  & SuperPoint~\cite{superpoint} & RoMa v1~\cite{romav1} & 3.64 & 6.63 & 12.91 & 0.26 & 1.42 & 7.21 \\ 
MP-SfM$^\star$~\cite{mpsfm} & D+I  & SuperPoint~\cite{superpoint} & RoMa v1~\cite{romav1} & 42.17 & 60.42 & 81.38 & 3.22 & 12.2 & \Third 40.29 \\ \cline{3-4}
DF-SfM~\cite{dfsfm} & I+R  & \multirow{4}{*}{RoMa v1~\cite{romav1}} & \multirow{4}{*}{RoMa v1~\cite{romav1}} & 79.32 & 88.42 & 95.82 & 3.13 & 9.79 & 29.10 \\ 
Dense-SfM~\cite{densesfm} & I+R  &  &  & \Second 84.79 & \Second 92.62 & \Second 97.77 & 7.38 & \Third 17.06 & 36.35 \\
MV-RoMa~\cite{mvroma} & I+R &  & & \Best 85.88 & \Best 92.99 & \Best 98.05 & 3.95 & 9.94 & 23.81 \\
\textbf{Ours} & D+G &  & & 60.12 & 74.90 & 90.17 & \Best 20.28 & \Best 47.53 & \Best 74.58 \\
\bottomrule
\end{tabular}
}
\end{table*}

\PAR{Point Cloud Accuracy.}
Following dense feature refinement methods~\cite{dfsfm, densesfm, mvroma}, we also evaluate the accuracy and completeness of the reconstructed point clouds.
We use the ETH3D training set, which contains 13 scenes with milimeter-accurate ground-truth point clouds.
Different from prior methods that uses the triangulated points using fixed camera poses and intrinsics, our method directly uses the optimized 3D points produced after bundle adjustment for evaluation.
We use the official ETH3D evaluation tool~\cite{eth3dsfm} and report accuracy and completeness at \SI{1}{\centi\metre}, \SI{2}{\centi\metre}, and \SI{5}{\centi\metre} thresholds.
Results are reported in Tab.~\ref{tab:eth3d_triang}.

With the same LoMa-B front-end, our method improves point accuracy over COLMAP and GLOMAP by a large margin, but results in lower completeness due to strict depth consistency filtering.
When using RoMa matches, our method produces substantially more complete reconstructions.
It achieves the best completeness across all thresholds, reaching $74.58$ at \SI{5}{\centi\metre}, outperforming feed-forward and dense reconstruction baselines.
However, specialized feature-refinement methods such as Dense-SfM and MV-RoMa achieve higher point accuracy by relying on additional multi-view matching or match refinement modules.
These methods are orthogonal to our contribution, which focuses on improving the SfM pipeline.
Since our method can directly benefit from stronger correspondences or refined tracks, such refinement modules could be readily integrated into our pipeline without any changes, and we expect performance to improve further.


\begin{table*}[t] \centering
\caption{Analysis of failure cases on 263 IMC-2021 scenes. For DGSfM, we ablate the impact of the depth-based modules, reporting 4 values per AUC column: using \cmark both / \xmark depth-filter \cmark pose init / \cmark depth-filter \xmark pose init / \xmark both.
}
\label{tab:failure_cases}
\resizebox{\linewidth}{!} {
\begin{tabular}{l|*{3}{c}}
\toprule
Method & AUC@3$\degree$ & AUC@5$\degree$ & AUC@10$\degree$ \\
\midrule
GLOMAP & 40.7 & 54.4 & 70.7 \\
GLOMAP - RePoseD w/o focal & 34.3 & 46.1 & 60.3 \\
GLOMAP - RePoseD w/ focal & 31.2 & 41.6 & 54.6 \\
\midrule \midrule
Ours w/ ZoeDepth & \colorbox{white}{21.1}/\colorbox{pastel_pink}{24.4}/\colorbox{pastel_blue}{25.2}/\colorbox{pastel_yellow}{25.4} & \colorbox{white}{31.6}/\colorbox{pastel_pink}{34.8}/\colorbox{pastel_blue}{35.7}/\colorbox{pastel_yellow}{35.8} & \colorbox{white}{47.0}/\colorbox{pastel_pink}{50.3}/\colorbox{pastel_blue}{51.1}/\colorbox{pastel_yellow}{51.3} \\
Ours w/ UniDepthV2 & \colorbox{pastel_yellow}{26.2}/\colorbox{pastel_blue}{26.1}/\colorbox{pastel_blue}{26.1}/\colorbox{pastel_pink}{26.0} & \colorbox{pastel_blue}{36.9}/\colorbox{pastel_yellow}{37.1}/\colorbox{pastel_pink}{36.7}/\colorbox{pastel_blue}{36.9} & \colorbox{pastel_pink}{52.2}/\colorbox{pastel_blue}{53.0}/\colorbox{white}{52.0}/\colorbox{pastel_yellow}{53.1} \\
Ours w/ MoGe1 & \colorbox{pastel_yellow}{30.0}/\colorbox{pastel_pink}{29.6}/\colorbox{pastel_blue}{29.8}/\colorbox{white}{29.3} & \colorbox{pastel_yellow}{42.0}/\colorbox{pastel_pink}{41.6}/\colorbox{pastel_blue}{41.9}/\colorbox{white}{41.4} & \colorbox{pastel_yellow}{58.9}/\colorbox{pastel_pink}{58.5}/\colorbox{pastel_blue}{58.7}/\colorbox{white}{58.3} \\
Ours w/ MoGe2 & \colorbox{pastel_yellow}{31.2}/\colorbox{pastel_blue}{31.0}/\colorbox{pastel_yellow}{31.2}/\colorbox{pastel_pink}{30.6} & \colorbox{pastel_blue}{43.4}/\colorbox{pastel_pink}{43.2}/\colorbox{pastel_yellow}{43.5}/\colorbox{white}{43.0} & \colorbox{pastel_blue}{60.8}/\colorbox{pastel_pink}{60.3}/\colorbox{pastel_yellow}{60.9}/\colorbox{white}{60.1} \\
\bottomrule
\end{tabular}
}
\end{table*}

\begin{table}[t] \centering
\caption{Ablation without (first 3 rows) and with (last 3 rows) Doppelgangers++ and triplet-guided edge pruning.}
\label{tab:graph_filtering}
\resizebox{0.8\linewidth}{!} {
\begin{tabular}{l|*{3}{c}|*{3}{c}}
\toprule
\multirow{2}{*}{Method} & \multicolumn{3}{c|}{ETH-3D~\cite{eth3dsfm}} & \multicolumn{3}{c}{IMC-2021~\cite{imc2021}} \\ 
& AUC@$1\degree$ & AUC@$3\degree$ & AUC@$5\degree$ & AUC@$3\degree$ & AUC@$5\degree$ & AUC@$10\degree$ \\
\midrule
COLMAP~\cite{colmap} & 43.45 & \Second 61.66 & \Second 66.32 & 42.40 & 54.27 & 67.36 \\
GLOMAP~\cite{glomap} & \Second 43.74 & 60.19 & 64.84 & \Second 45.91 & \Second 58.29 & \Second 71.81 \\
\textbf{Ours} & \Best 45.88 & \Best 69.87 & \Best 77.49 & \Best 46.54 & \Best 60.12 & \Best 73.95 \\
\midrule
\midrule
COLMAP~\cite{colmap} & 46.53 & 69.20 & 74.22 & 43.84 & 55.59 & 68.24 \\
GLOMAP~\cite{glomap} & \Second 48.63 & \Second 70.20 & \Second 75.10 & \Second 46.29 & \Second 58.60 & \Second 72.03 \\
\textbf{Ours} & \Best 51.60 & \Best 78.25 & \Best 85.43 & \Best 46.59 & \Best 59.91 & \Best 75.59 \\
\bottomrule
\end{tabular}
}
\end{table}

\begin{table*}[t] \centering
\caption{Timing breakdown (in sec.) on ETH3D (averaged).}
\label{tab:timing}
\resizebox{\linewidth}{!} {
\begin{tabular}{l|c|c|c|c|c}
\toprule
Module & MonoDepth & Feat. matching (LoMa-B) & Dopp++~\cite{doppelgangers++} \& Triplet~\cite{manam2024leveraging} & Mapping & Total  \\
\midrule
GLOMAP~\cite{glomap} & - & \multirow{2}{*}{17.42} & \multirow{2}{*}{24.89} & 16.59 & 58.90 \\
Ours & 2.87 &  &  & 16.49 & 61.67 \\
\bottomrule
\end{tabular}
}
\end{table*}

\begin{table*}[t] \centering
\caption{Ablation of individual constraints on ETH3D.}
\label{tab:corr_filtering}
\resizebox{0.6\linewidth}{!} {
\begin{tabular}{l|*{3}{c}}
\toprule
Method & AUC@$1\degree$ & AUC@$3\degree$ & AUC@$5\degree$ \\
\midrule
Reprojection error only & \Second 48.83 & \Second 75.02 & \Second 82.70 \\
Depth consistency only & 46.20 & 74.64 & 82.48 \\
Both & \Best 51.60 & \Best 78.25 & \Best 85.43 \\
\bottomrule
\end{tabular}
}
\end{table*}

\PAR{Potential Failure Cases.} 
Sec. 6 in the main paper discusses potential failures, in particular failures caused by relying on depth estimates.  
To better understand this reliance, we identified 263 scenes from IMC-2021 (out of all 1,559 scenes) where the AUC scores of DGSfM (using MoGe2~\cite{wang2025moge2}) are 20\% or more lower than GLOMAP's AUC scores. 
Tab.~\ref{tab:failure_cases} compares our approach (with multiple MDEs, including the outdated ZoeDepth~\cite{zoedepth}) against GLOMAP on these scenes. 
Disabling the other stages relying on depth estimates (depth filter,  pose-point initialization) shows that these stages have little impact on the performance for modern depth estimators (UniDepthV2~\cite{piccinelli2025unidepthv2}, MoGe1~\cite{wang2025moge}/2~\cite{wang2025moge2}). 
Only when using ZoeDepth do we observe a gain when disabling our proposed filtering stages. 
This suggests that the limiting factor is the accuracy of the relative poses and intrinsics estimated using RePoseD~\cite{ding2025reposed}. 
To verify this hypothesis, we applied GLOMAP on the same set of relative poses (RePoseD with MoGe2). 
As can be seen, its performance decreases substantially, with a further decline when RePoseD's estimated focal lengths are also used. 
Visual inspection of the scenes shows that the images are typically taken far away from the scene, \ie, these are scenes for which we expect depth estimates to be inaccurate. 
An interesting direction for future work is to automatically detect when predicted depth are inaccurate and to fall back on pure image-based reasoning in these cases. 

Overall, DGSfM is not universal. 
It is intended for unordered image collections in which a sufficiently connected view graph and a reasonably informative monocular depth prior can be obtained.
As discussed above, DGSfM can struggle/fail when depth predictions are coherently wrong or image overlap is too weak to maintain graph connectivity. 

\PAR{Contribution of Non-depth Modules.} 
In Tab.~\ref{tab:graph_filtering}, we compare DGSfM with COLMAP and GLOMAP without and with the use of Doppelgangers++ and triplet-guided edge pruning. 
DGSfM outperforms both in both cases (GLOMAP performs close  on IMC-2021 for AUC@$3\degree$).

\PAR{Per-module Timing.}
Tab.~\ref{tab:timing} reports the full runtime and per-module breakdown of our method using MoGe2~\cite{wang2025moge2} compared to GLOMAP using the same LoMa-B~\cite{nordstrom2026loma} matches.

\PAR{Ablation of individual constraints in correspondence filtering.}
Tab.~\ref{tab:corr_filtering} ablates the two individual constraints (reprojection error and depth consistency). 
Both contribute to the overall performance, with repojection error providing (slightly) stronger constraints. 
While ``depth filtering may [indeed] accept internally consistent erroneous solutions", we typically observe that using depths improves performance. 
Even if the depths are inaccurate, we do not observe filtering reducing pose accuracy (\cf Tab.~\ref{tab:failure_cases}).

\begin{figure*}[t]
  \centering
\includegraphics[width=\textwidth]{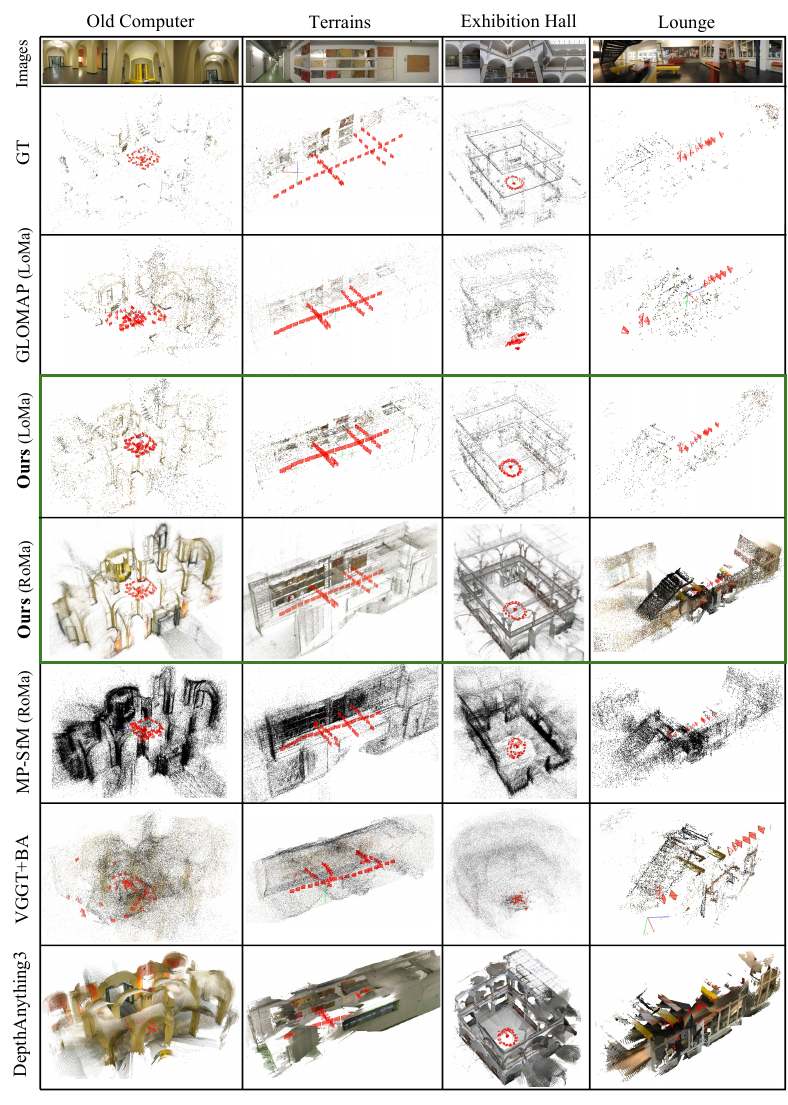}
  \caption{Qualitative comparison on ETH3D~\cite{eth3dsfm} indoor scenes.
  }
  \label{fig:eth3d_indoor}
\end{figure*}

\begin{figure*}[t]
  \centering
\includegraphics[width=\textwidth]{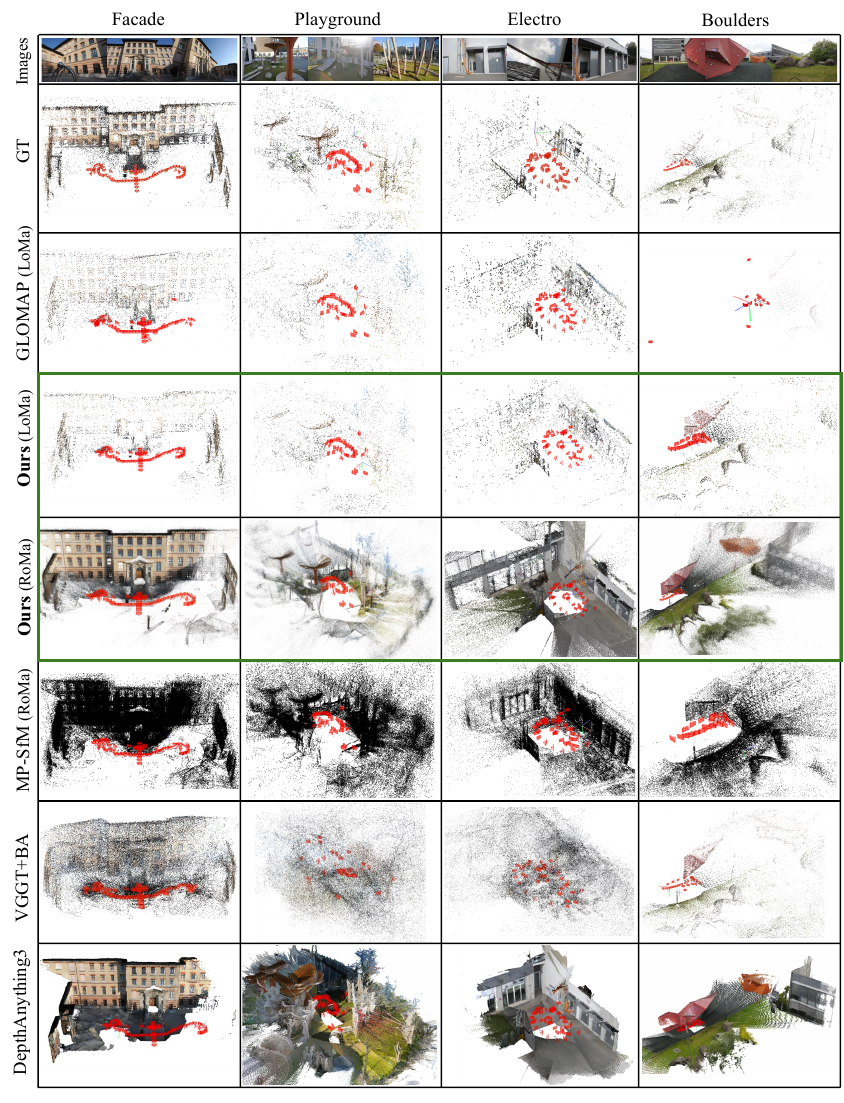}
  \caption{Qualitative comparison on ETH3D~\cite{eth3dsfm} outdoor scenes.
  }
  \label{fig:eth3d_outdoor}
\end{figure*}

\begin{figure*}[t]
  \centering
\includegraphics[width=\textwidth]{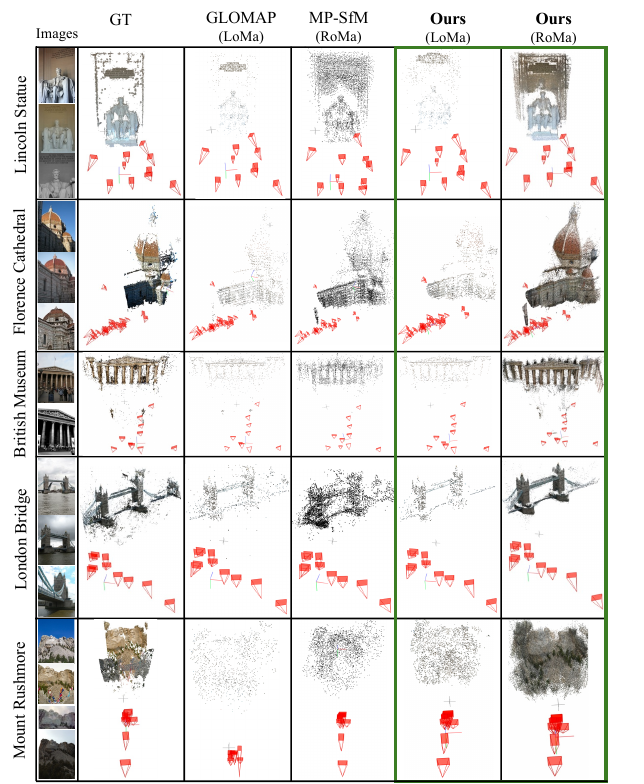}
  \caption{Qualitative comparison on IMC-2021~\cite{imc2021} scenes with different image bags.
  }
  \label{fig:imc21_results}
\end{figure*}

\section{Additional Visualization}
\label{sec:add_vis}

\PAR{ETH3D.}
Figures \ref{fig:eth3d_indoor} and \ref{fig:eth3d_outdoor} provide qualitative comparisons on indoor and outdoor scenes from ETH3D~\cite{eth3dsfm}.
We compare the ground-truth reconstruction with GLOMAP~\cite{glomap}, our method using LoMa~\cite{nordstrom2026loma} matches, our method using RoMa v1~\cite{romav1} matches, MP-SfM~\cite{mpsfm}, VGGT+BA~\cite{vggt}, and DepthAnything3~\cite{lin2025depthanything3}.

With the same LoMa matches, our method produces more accurate camera poses than GLOMAP but fewer points, which are cleaner and more accurate, as also supported by the triangulation results in Tab.~\ref{tab:eth3d_triang}.
When using dense RoMa matches, our method reconstructs substantially more complete 3D structure while maintaining stable camera poses, demonstrating that the proposed pipeline can benefit from dense correspondences without relying on additional dense feature refinement modules.

Compared with MP-SfM, our method obtains cleaner point clouds and more consistent camera poses across the evaluated scenes.
Feed-forward methods such as DepthAnything3 produce denser geometry, but their camera poses are far less accurate and the reconstructed point clouds can be misaligned (\textit{Playground} scene) across views.
Overall, these qualitative results show that our approach provides a favorable balance between pose accuracy, point-cloud cleanliness, and reconstruction completeness.

\PAR{IMC2021.}
Figure.~\ref{fig:imc21_results} shows qualitative comparisons on IMC2021~\cite{imc2021} scenes with different image bags.
IMC2021 provides pseudo ground-truth camera poses obtained from full image sets of COLMAP~\cite{colmap, schoenberger2016mvs} MVS reconstructions, which typically contain denser scene geometry than sensor-measured ground-truth in ETH3D.
We observe trends similar to ETH3D; with the same LoMa matches, our method produces slightly more accurate camera poses and cleaner sparse point tracks than GLOMAP.
Compared with MP-SfM, our method yields more stable camera poses on scenes such as \textit{Lincoln Statue} and \textit{British Museum}, where MP-SfM produces noisier and less complete point clouds.
Our method with RoMa matches further improves reconstruction completeness with substantially richer 3D structure from a small number of cameras while maintaining globally consistent camera poses.

\section*{Acknowledgements}
This work was supported by the Czech Science Foundation (GACR) JUNIOR STAR, Grant No.~22-23183M. The access to the computational infrastructure of the OP VVV funded project CZ.02.1.01/0.0/0.0/16\_019/0000765 ``Research Center for Informatics'' is also gratefully acknowledged. V.K. was funded by the EU NextGenerationEU through the Recovery and Resilience Plan for Slovakia under the project No. 09I02-03-V01-00029. 

\newpage
\clearpage

\bibliographystyle{splncs04}
\bibliography{main}
\end{document}